\documentclass[aps,pra,reprint,showkeys,amssymb,floatfix,longbibliography,superscriptaddress]{revtex4-1}

\usepackage{graphicx}
\usepackage{amsmath,amsfonts,amssymb}
\usepackage{hyperref}
\usepackage{braket}
\usepackage{xcolor}
\usepackage{tikz}
\usepackage{color}
\usepackage{enumerate}
\usepackage[caption=false]{subfig}
\usepackage{multirow}
\usepackage{qcircuit}
\usepackage[normalem]{ulem}
\usepackage{placeins}
\usepackage{url}
\usepackage{bbm}

\newcommand{\appref}[1]{Appendix~\ref{#1}}
\newcommand{\secref}[1]{Sec.~\ref{#1}}
\newcommand{\ssecref}[1]{Section~\ref{#1}}
\newcommand{\figref}[1]{Fig.~\ref{#1}}
\newcommand{\tabref}[1]{Table~\ref{#1}}
\newcommand{\equref}[1]{Eq.~\eqref{#1}}

\newcommand{\equsref}[2]{Eqs.~\eqref{#1}--\eqref{#2}}

\newcommand\boxit[1]{\begin{center}\fbox{\parbox{.8\linewidth}{#1}}\end{center}}

\begin{document}

\title{Support vector machines on the D-Wave quantum annealer}

\author{D. Willsch}
\affiliation{Institute for Advanced Simulation,
  J\"ulich Supercomputing Centre,\\
  Forschungszentrum J\"ulich, D-52425 J\"ulich, Germany}
\affiliation{RWTH Aachen University, D-52056 Aachen, Germany}
\author{M. Willsch}
\affiliation{Institute for Advanced Simulation,
  J\"ulich Supercomputing Centre,\\
  Forschungszentrum J\"ulich, D-52425 J\"ulich, Germany}
\affiliation{RWTH Aachen University, D-52056 Aachen, Germany}
\author{H. De Raedt}
\affiliation{Zernike Institute for Advanced Materials,\\
University of Groningen, Nijenborgh 4, NL-9747 AG Groningen, The Netherlands}
\author{K. Michielsen}
\affiliation{Institute for Advanced Simulation,
  J\"ulich Supercomputing Centre,\\
  Forschungszentrum J\"ulich, D-52425 J\"ulich, Germany}
\affiliation{RWTH Aachen University, D-52056 Aachen, Germany}

\date{\today}

\begin{abstract}

  Kernel-based support vector machines (SVMs) are supervised machine learning
  algorithms for classification and regression problems. We introduce a method to
  train SVMs on a D-Wave 2000Q quantum annealer and study its performance in
  comparison to SVMs trained on conventional computers. The method is applied to
  both synthetic data and real data obtained from biology experiments. We find
  that the quantum annealer produces an ensemble of different solutions that
  often generalizes better to unseen data than the single global minimum of an
  SVM trained on a conventional computer, especially in cases where only limited
  training data is available. For cases with more training data than currently
  fits on the quantum annealer, we show that a combination of classifiers for
  subsets of the data almost always produces stronger joint classifiers than the
  conventional SVM for the same parameters.

\end{abstract}


\maketitle

\section{Introduction}

The growing interest in both quantum computing and machine learning has inspired
researchers to study a combination of both fields, termed \emph{quantum machine
learning} \cite{Neven2008BinaryClassificationOnDWave, Pudenz2012QML,
Neven2012QBoost, Adachi2015QMLApplicationToDeepLearning,
Potok2017DWMLComplexDLNetworks, OMalley2017DWMLNNBinaryMatrixFactorization,
Ottaviani2018DWMLLowRankMatrixFactorization}. Recently, it has been shown that
using the D-Wave quantum annealer can yield advantages in classification
performance over state-of-the-art conventional approaches for certain
computational biology problems using a linear classifier
\cite{Li2018TFDNABindingAffinity}. In this paper, we improve on these results by
replacing the linear classifier with a superior nonlinear classification
approach, the kernel-based support vector machine (SVM)
\cite{schoelkopf2002learningwithkernels, Bishop2006MachineLearning}. We
introduce its formulation for a D-Wave quantum annealer and present training
results for both synthetic data and real data. To distinguish between the SVM
formulations, we use the word \emph{classical} to denote the original version of
an SVM as defined in \cite{schoelkopf2002learningwithkernels}.

The field of supervised machine learning deals with the problem of learning
model parameters from a set of labeled training data in order to make
predictions about test data. SVMs in particular are known for their stability
(in comparison to decision trees or deep neural networks
\cite{Li2002DecisionTreeInstability,  Yuan2003DecisionTreesVSSVMs,
Xu2009RobustnessSVM, Raczko2017SVMvsRandomForestandNN}), in the sense that small
differences in the training data do not generally produce huge differences in
the resulting classifiers. Moreover, kernel-based SVMs profit from the
\emph{kernel trick}, effectively  maneuvering around the ``curse of
dimensionality'' \cite{schoelkopf2002learningwithkernels,numericalrecipes}.
In contrast to Deep Learning, which often requires large amounts of training
data, SVMs are typically used when only small sets of training data are
available. But also in combination with Deep Learning, where SVMs are applied on
top of neural networks to classify the detected features, SVMs have been
found to yield significant gains in classification performance
\cite{Tang2013DLwithSVM, Kim2013DeepNetworkWithSVM, Lazri2018SVMwithANNandRF,
Zareapoor2018KernelSVMwithDL}.

Quantum annealers manufactured by D-Wave Systems Inc.~are available with about
2000 qubits \cite{Harris2010DWave, Johnson2011DWave, Bunyk2014DWave, DW2000Q}.
They automatically produce a variety of close-to-optimal solutions to a given
optimization problem \cite{Mott2017HiggsQML, Li2018TFDNABindingAffinity,
DW2000Q}. This is particularly interesting in the context of machine learning,
because any of the solutions produced for a given \emph{training dataset} have
the potential to perform well on new \emph{test data}. For SVMs, for which the
original solution is the \emph{global optimum} of the underlying convex
optimization problem for the training data \cite{Bishop2006MachineLearning}, it
is an interesting question whether the ensemble of different solutions from the
quantum annealer can improve the classification performance for the test data.

We conduct our SVM experiments on a D-Wave 2000Q (DW2000Q) quantum annealer
\cite{DW2000Q}. Quantum annealing (QA) is so far the only paradigm of quantum
computing for which processors of a reasonable size are available. The other
paradigm of quantum computing, i.e., the gate-based (or universal) quantum
computer \cite{NielsenChuang}, is still limited to less than 100 quantum bits
(qubits) \cite{NationalAcademyOfSciences2018QuantumComputing}. It is worth
mentioning that for gate-based quantum computers, a quantum algorithm for SVMs
has already been proposed \cite{Rebentrost2014gatebasedQSVMtheory}. However,
only a few very simple tasks, for which almost all classification was already
done in the preprocessing step, have been studied experimentally
\cite{Li2015gatebasedQSVMexperiment}.

QA requires the formulation of the computational problem as a quadratic
unconstrained binary optimization (QUBO). A QUBO problem is defined as the
minimization of the energy function
\begin{align}
  \label{eq:qubo}
  E = \sum_{i\le j} a_i Q_{ij} a_j,
\end{align}
where $a_i\in\{0,1\}$ are the binary variables of the optimization problem, and
$Q$ is an upper-triangular matrix of real numbers called the QUBO weight matrix.
Note that the size of the DW2000Q quantum processor and the Chimera topology
\cite{Bunyk2014DWave} impose certain restrictions on this matrix. A popular
alternative formulation of the problem in terms of variables $s_i\in\{-1,1\}$ is
known as the Ising model \cite{Ising1925, Barahona1982IsingSpinGlassModel}.

We present a formulation of SVMs as a QUBO defined by \equref{eq:qubo} and
discuss certain mathematical properties in the training of SVMs that make it
particularly appealing for use on a quantum annealer. In comparison to the
classical SVM, we find that a combination of the solutions returned by the
quantum annealer often surpasses the single solution of the classical SVM.

This paper is structured as follows: In \secref{sec:SVM}, we introduce the
classical SVM, our formulation of an SVM for QA, and the metrics we use  to
compare the performance of both. \ssecref{sec:applications} contains the
application of both SVM versions to synthetic two-dimensional data and real
data from biology experiments, including the calibration, training, and testing
phase. We conclude our study with a short discussion in \secref{sec:conclusion}.

\section{SVMs on a quantum annealer}
\label{sec:SVM}

In this section, we first briefly review the classical SVM, and then introduce
the QA version of an SVM. Finally, we discuss ways to evaluate the
classification performance in the applications presented in the next section.

\subsection{The classical SVM}
\label{sec:cSVM}

An SVM is a supervised machine-learning algorithm for classification and
regression. It operates on a dataset
\begin{align}
  \label{eq:data}
  D = \{ (\mathbf x_n, t_n) : n = 0,\ldots,N-1 \},
\end{align}
where $\mathbf x_n \in \mathbb R^d$ is a point in $d$-dimensional  space (a
\emph{feature vector}), and $t_n$ is the target label assigned to $\mathbf x_n$.
We consider the task of learning a binary classifier that assigns a class label
$t_n = \pm 1$ for a given data point $\mathbf x_n$. In the following, we call
the class $t_n = 1$ ``positive'' and the class $t_n = -1$ ``negative''.

Training an SVM amounts to solving the quadratic programming (QP)
problem \cite{numericalrecipes}
\begin{align}
   &\text{minimize} && E = \frac 1 2 \sum\limits_{nm} \alpha_n \alpha_m t_n t_m k(\mathbf x_n, \mathbf x_m)\nonumber\\
  \label{eq:qp}
    &&&\qquad  - \sum\limits_n \alpha_n,\\
  \label{eq:qpb}
  &\text{subject to} && 0 \le \alpha_n \le C, & \\
  \label{eq:qpzero}
  &\text{and} && \sum\limits_n \alpha_n t_n = 0,
\end{align}
for $N$ coefficients $\alpha_n\in\mathbb R$, where $C$ is a regularization
parameter and $k(\cdot,\cdot)$ is the kernel function of the SVM
\cite{schoelkopf2002learningwithkernels}. The resulting coefficients $\alpha_n$
define a $(d-1)$-dimensional decision boundary that separates $\mathbb R^d$ in
two regions corresponding to the predicted class label.
A typical solution often contains many $\alpha_n=0$. The decision boundary
is then determined by the points corresponding to $\alpha_n\neq0$ (the
\emph{support vectors} of the SVM).
A prediction for an
arbitrary point $\mathbf x\in\mathbb R^d$ can be made by evaluating the decision
function
\begin{align}
  \label{eq:decisionfunction}
  f(\mathbf x) = \sum\limits_n \alpha_n t_n k(\mathbf x_n,\mathbf x) + b,
\end{align}
where a reasonable choice to determine the bias $b$ is given by
\cite{numericalrecipes}
\begin{align}
  \label{eq:biasaveraging}
  b = \frac{\sum\limits_n \alpha_n (C-\alpha_n) \left[t_n - \sum\limits_m \alpha_m t_m k(\mathbf x_m,\mathbf x_n)\right]}
  {\sum\limits_n \alpha_n (C-\alpha_n)}.
\end{align}

Geometrically, the decision function $f(\mathbf x)$ represents a signed distance
between the point $\mathbf x$ and the decision boundary. Thus the class label
for $\mathbf x$ predicted by the trained SVM is $\widetilde{t} =
\mathrm{sign}(f(\mathbf x))$.

The formulation of the problem given in \equsref{eq:qp}{eq:qpzero} is the
so-called dual formulation of an SVM (see \cite{Bishop2006MachineLearning} for
more information). Since it represents a convex quadratic optimization problem,
it is one of the rare minimization problems in machine learning that have a
global minimum. Note, however, that the global minimum with respect to the
training dataset $D$ may not necessarily be optimal for generalizing to the test
dataset.

Kernel-based SVMs are particularly powerful since they allow for nonlinear
decision boundaries defined by $f(\mathbf x)=0$ (see
\equref{eq:decisionfunction}),  implicitly mapping the feature vectors to
higher-dimensional spaces \cite{Burges1998SVMTutorial}. Interestingly, the
complexity of the problem does not grow with this dimension, since only the
values of the kernel function $k(\mathbf x_n, \mathbf x_m)$ enter the problem
specification (see \equref{eq:qp}). This fact is known as the \emph{kernel
trick} \cite{schoelkopf2002learningwithkernels,numericalrecipes}.

The choice of the kernel function can have a significant impact on the results.
Typical choices for SVMs are linear, polynomial, sigmoid, and radial basis function (rbf) kernels
\cite{Bishop2006MachineLearning}. In general, an rbf kernel is a kernel for which
$k(\mathbf x_n, \mathbf x_m)$ can be written as a function of the distance $\|\mathbf x_n - \mathbf x_m\|$ only
\cite{schoelkopf2002learningwithkernels}. The most common rbf kernel is the Gaussian
kernel (often referred to as \emph{the} rbf kernel),
\begin{align}
  \label{eq:rbf}
  \mathrm{rbf}(\mathbf x_n, \mathbf x_m) = e^{-\gamma\|\mathbf x_n - \mathbf x_m\|^2},
\end{align}
where the value of the hyperparameter $\gamma>0$ is usually determined in a
calibration procedure prior to the training phase (if no particular set of
values for $\gamma$ is known for the data, a good strategy is to try
exponentially growing sequences like $\gamma \in \{ \ldots, 2^{-3}, 2^{-2},
\ldots \}$ \cite{Hsu1992PracticalGuideSVM}).

Gaussian kernels have the advantage of not suffering as much from numerical difficulties as polynomial kernels
\cite{Hsu1992PracticalGuideSVM} and, in general, compare favorably to sigmoid
or tanh kernels (which are, strictly speaking, not positive semi-definite)
\cite{Lin2003SVMsigmoidkernel}. They implicitly map the feature vector onto an
infinite-dimensional space \cite{Bishop2006MachineLearning}. In principle, a
Gaussian kernel also includes the linear kernel as an asymptotic case
\cite{Keerthi2003SVMasymptoticRBFkernel}. However, we explicitly include a
linear kernel for convenience, denoted by the special value $\gamma=-1$.
Therefore, we formally define
\begin{align}
  \label{eq:kernel}
  k(\mathbf x_n, \mathbf x_m) := \begin{cases}
    \mathrm{rbf}(\mathbf x_n, \mathbf x_m) & (\gamma > 0) \\
    \mathbf x_n \cdot \mathbf x_m & (\gamma = -1),
  \end{cases}
\end{align}
as the kernel function for our experiments.

\boxit{
  In the following, we symbolically write $\texttt{cSVM}(C,\gamma)$ to denote the
  training of the classical SVM defined by \equsref{eq:qp}{eq:qpzero} with the kernel
  function given in \equref{eq:kernel}.
}
For the computational work associated with \texttt{cSVM}, we used the C++
library LIBSVM \cite{Chang2001LibSVM}, the Python module Scikit-learn
\cite{scikitlearn}, and a quadratic programming solver from the Python package
CVXOPT \cite{cvxopt}. All packages produced identical results, i.e.,
the global optimum of the convex optimization problem.

\subsection{The quantum SVM}
\label{sec:qSVM}

The solution to \equsref{eq:qp}{eq:qpzero} consists of real numbers
$\alpha_n\in\mathbb R$. However, the DW2000Q can only produce discrete, binary
solutions to a QUBO (see \equref{eq:qubo}). Therefore, we use an encoding of the
form
\begin{align}
  \label{eq:encoding}
  \alpha_n = \sum\limits_{k=0}^{K-1} B^k a_{Kn+k},
\end{align}
where $a_{Kn+k} \in\{0,1\}$ are binary variables, $K$ is the number of binary
variables to encode $\alpha_n$, and $B$ is the base used for the encoding. In
practice, we obtained good results for $B=2$ or $B=10$ and a small number of
$K$ (see also the list of arguments given below).

To formulate the QP problem given in \equsref{eq:qp}{eq:qpzero} as a QUBO (see
\equref{eq:qubo}), we use the encoding defined in \equref{eq:encoding} and
introduce a  multiplier $\xi$ to include the second constraint given in
\equref{eq:qpzero} as a squared penalty term. We obtain
\begin{align}
  E &= \frac 1 2 \sum\limits_{nmkj} a_{Kn+k} a_{Km+j} B^{k+j} t_n t_m k(\mathbf x_n, \mathbf x_m)\nonumber\\
  \label{eq:qsvmfull}
    &\quad - \sum\limits_{nk} B^k a_{Kn+k} + \frac{\xi}{2} \left(\sum\limits_{nk} B^k a_{Kn+k} t_n\right)^2 \\
    &= \sum\limits_{n,m=0}^{N-1} \sum\limits_{k,j=0}^{K-1} a_{Kn+k}
     \widetilde{Q}_{Kn+k,Km+j} a_{Km+j},
\end{align}
where $\widetilde{Q}$ is a matrix of size $KN\times KN$ given by
\begin{align}
  \widetilde{Q}_{Kn+k,Km+j} &= \frac 1 2 B^{k+j} t_n t_m (k(\mathbf x_n, \mathbf x_m)+\xi) \nonumber\\
  \label{eq:qsvmqubo}
  &- \delta_{nm} \delta_{kj} B^k.
\end{align}
Since $\widetilde{Q}$ is symmetric, the upper-triangular QUBO matrix $Q$
required for the QUBO formulation given in \equref{eq:qubo} is defined by
$Q_{ij}=\widetilde{Q}_{ij}+\widetilde{Q}_{ji}$ for $i<j$ and $Q_{ii} =
\widetilde{Q}_{ii}$.
Note that the constraint \equref{eq:qpb} is automatically included in
\equref{eq:qsvmfull} through the encoding given in \equref{eq:encoding}, since
the maximum for $\alpha_n$ is given by
\begin{align}
  \label{eq:boxconstraintparameter}
  C = \sum\limits_{k=1}^K B^k,
\end{align}
and $\alpha_n\ge0$ by definition.

Given $K$, each $\alpha_n$ can take only $2^K$ different values according to
\equref{eq:encoding}. At first, it may seem questionable why a small
number of $B$ and $K$ should be sufficient. The following arguments and
empirical findings for SVMs motivated us to try the QUBO approach:

\begin{enumerate}

  \item A typical solution to \equsref{eq:qp}{eq:qpzero} consists of many
  $\alpha_n = 0$ with only a few $\alpha_m \neq 0$ (the corresponding data
  points $\{\mathbf x_m\}$ are the support vectors). On a digital computer using
  floating-point numbers, establishing convergence to exactly $0$ is a subtle
  task, whereas the encoding in \equref{eq:encoding} directly includes this
  value.

  \item The box constraint \equref{eq:qpb} is automatically satisfied by the
  choice of the encoding \equref{eq:encoding} (see
  \equref{eq:boxconstraintparameter}).

  \item In principle, one can extend the encoding \equref{eq:encoding} to
  fractional numbers by replacing the base $B^k$ with $B^{k-k_0}$ for some
  $k_0\in\mathbb N$. Eventually, this would yield the same range of
  floating-point numbers as used in conventional digital computers, namely the
  IEEE standard for floating-point arithmetic \cite{ieee754std}. However,
  it was observed that this kind of precision is not required for SVMs to
  produce reasonable results (see
  \cite{Lesser2011SVMwithReducedFloatPrecision}), and it would also not be
  feasible with the current generation of QA devices.

  \item For the classification task addressed by an SVM, the global order of
  magnitude of all $\alpha_n$ is often not as important as the relative factors
  between different $\alpha_n$. This can be understood by studying the effect of
  substituting $\alpha_n\mapsto S\alpha_n$ for some factor $S$ in
  \equsref{eq:qp}{eq:qpzero}. Since $E$ and $E/S^2$ are optimal for the same
  $\{\alpha_n\}$, and the hyperparameters of the box constraint are calibrated
  separately, it only replaces the linear term in \equref{eq:qp} by $-\sum_n
  \alpha_n/S$. This term only affects the size of the margin between the
  decision boundary and the support vectors (see also
  \cite{Bishop2006MachineLearning}). However, if this is still found to be an
  issue, one can simply adjust the encoding \equref{eq:encoding} accordingly.

  \item Especially for the Gaussian kernel given in \equref{eq:rbf}, points with
  a large distance $\|\mathbf x_n - \mathbf x_m\|\gg1$ result in $k(\mathbf
  x_n,\mathbf x_m)\approx0$. This can be used to reduce couplings between the
  qubits such that embedding the problem on the quantum annealer is less
  complex. This may either yield better solutions or allow larger problems to be
  embedded on the DW2000Q.

  \item The constraint $\sum_n \alpha_n t_n=0$ mathematically corresponds to an
  optimal bias $b$ in the decision function given in
  \equref{eq:decisionfunction} (see \cite{Bishop2006MachineLearning}). We have
  included it in \equref{eq:qsvmfull} through the multiplier $\xi$. However, the
  constraint need not be satisfied exactly for the classification task to
  produce good results. Since the bias $b$ is only one parameter, it can easily
  be adjusted afterwards if necessary. For this reason, it can be that $\xi=0$
  already suffices to get reasonable  results. Furthermore, the special value
  $\xi=1$ yields the Mangasarian--Musicant variant of an SVM (see
  \cite{MangasarianMusicant1999SVMoverrelaxation,
  MangasarianMusicant2001NonlinearGeneralizedSVM} for more information). This
  variant has been shown to produce equally good classifiers while 
  being numerically much more tractable without requiring $\sum_n \alpha_n t_n=0$ \cite{numericalrecipes}. An
  alternative approach would be to include $\xi$ in the parameter set that has
  to be optimized (as conventionally done for Lagrange multipliers) by choosing
  an additional encoding for $\xi$ such as \equref{eq:encoding}. In this case,
  it would suffice  to replace the last term in \equref{eq:qsvmfull} by the
  linear penalty term $\xi\sum_n \alpha_n t_n$.  We experimented with this approach
  and it yields similar but less robust results (data not shown). For this
  reason, and due to the (on present quantum annealers) small set of numbers
  represented by the encoding \equref{eq:encoding}, and also because of the
  SVM's sensitivity to the bias, we found it more convenient to keep $\xi$ as a
  hyperparameter, and if necessary adjust the bias afterwards (see also
  \appref{app:bias}).

\end{enumerate}

The last step required to run the optimization problem on the DW2000Q is the
embedding  procedure \cite{Choi2008Embedding, Cai2014DWaveEmbedding}. It is
necessary because in general, the QUBO given in \equref{eq:qubo} includes some
couplers $Q_{ij} \neq 0$ between qubit $i$ and qubit $j$ for which no physical
connection  exists on the chip (the connectivity of the DW2000Q is given by the
Chimera topology \cite{Bunyk2014DWave}). The idea of embedding is to combine
several  physical qubits to one logical qubit (also called \emph{chain}) by
choosing a large negative value for their coupling strengths to favor solutions
where  the physical qubits are aligned. This can be used to increase the logical
connectivity between the qubits.

We use a function provided by D-Wave Systems Inc.~to generate embeddings for the
QUBOs given by \equref{eq:qsvmqubo} \cite{DWOceanSDK}. When no embedding can be
found, we successively decrease the number of  nonzero couplers
$n_{\mathrm{cpl}}$ by setting the smallest couplers to zero until  an embedding
is found. This works especially well in combination with the Gaussian  kernel
given in \equref{eq:rbf}, where points with a large squared distance  $\|\mathbf
x_n - \mathbf x_m\|^2$ only produce negligible contributions to the QUBO.
Typical values for $n_{\mathrm{cpl}}$ for the applications discussed in
\secref{sec:applications} are  between 1600 and 2500, while the number of
required qubits ranges from 28 to 114 with peaks at 56, 58, 84, and 87.

We chose to test the default mode of operation of the DW2000Q with an annealing
time of $20\,\mu\mathrm{s}$ and leave the analysis of improving the QA results
by advanced features like reverse annealing, spin-reversal
transforms, special annealing schedules, or alternative embedding
heuristics to the future \cite{Ohkuwa2018DWaveReverseAnnealing,
Boixo2013ExperimentalSignatureQA, DW2000Q}.

To summarize, the final QA version of the SVM defined by the QUBO in
\equref{eq:qsvmqubo} depends on the following hyperparameters: the encoding base
$B$, the number $K$ of qubits per coefficient $\alpha_n$, the
multiplier $\xi$, and the kernel parameter $\gamma$ (the number
$n_{\mathrm{cpl}}$ of strongest couplers embedded on the DW2000Q is different
for every run and is not a parameter of the SVM itself).
\boxit{
We denote the QA version of an SVM defined in \equref{eq:qsvmqubo} as
$\texttt{qSVM}(B,K,\xi,\gamma)$,
by analogy with  $\texttt{cSVM}(C,\gamma)$ defined
in \equsref{eq:qp}{eq:qpzero},
}
For each run on the DW2000Q, we consider the twenty lowest-energy samples from
10,000 reads, denoted by $\texttt{qSVM}(B,K,\xi,\gamma)\#i$ for $i=0,\ldots,19$.
Note that the cut at $i=20$ is arbitrary; one could also consider 50 or more
samples from the distribution if appropriate.

In principle, it can happen that a particular sample $\#i$ yields only
$\alpha_n=0$ or $\alpha_n=C$ such that the bias $b$  in
\equref{eq:biasaveraging} is undefined. This reflects the rare situation that no
support vectors have been found. In this case, one may simply discard the
affected sample and consider only the remaining samples.

Note that the DW2000Q produces a variety of close-to-optimal solutions (i.e., a
variety of different coefficients $\{\alpha_n\}^{(i)}$ obtained from
\equref{eq:encoding}). Many of these solutions may have a slightly higher energy
than the global minimum $\{\alpha_n\}^*$ found by \texttt{cSVM}, but still solve
the classification problem for the training data as intended. The different
solutions often emphasize different features of the training data. When applied
to the test data, a combination of these solutions has the potential to solve
the classification task better than \texttt{cSVM}, which only yields the global
minimum for the training data.

For the computational work associated with \texttt{qSVM}, we used the D-Wave
Ocean SDK \cite{DWOceanSDK}, which provides the functionality to generate
embeddings (see above) and produce results for the QUBO matrix defined in \equref{eq:qsvmqubo}.

\subsection{Using accuracy, AUROC, and AUPRC to assess the classification performance}
\label{sec:evaluation}

To measure the classification performance, we consider a separation of the data
$D$ given in \equref{eq:data} into two disjoint subsets $D^{(\mathrm{train})}$
and $D^{(\mathrm{test})}$. The training data $D^{(\mathrm{train})}$ is used to
train either $\texttt{cSVM}(C,\gamma)$ or $\texttt{qSVM}(B,K,\xi,\gamma)$. In
both cases, the result of the training is the set of coefficients
$\{\alpha_n\}$, which can be used to make class predictions by means of the
decision function given in \equref{eq:decisionfunction}. The classifier is then
evaluated for the test data $D^{(\mathrm{test})}$ by comparing the class
prediction $\widetilde{t_n} = \mathrm{sign}(f(\mathbf x_n))$ with the true label
$t_n$ for each $(\mathbf x_n,t_n)\in D^{(\mathrm{test})}$ from the test data.

A straightforward method to assess the performance of a classifier is to count
the number of correct predictions, i.e., the number of true positives
$\mathrm{TP}$ for which $\tilde t_n = t_n = 1$. Dividing this number by the
total number of points $|D^{(\mathrm{test})}|$ yields the \emph{classification
accuracy}. However, in binary classification problems, the accuracy is generally
considered a bad measure \cite{Provost1998AccuracyBadMeasure,
Cortes2003AUROCvsErrorRate}, because a higher accuracy does not necessarily
imply that the classifier is better. As a simple example, consider a dataset
with 80\% negatives. A trivial all-negative classifier, which always returns
$-1$, would already achieve an accuracy of 80\%, even though it is practically
useless. Instead, we are often interested in identifying good positives,
especially if the dataset contains a lot of negatives.

To obtain a more robust measure, we first count the number of all
cases that can occur when making the class prediction
$\widetilde{t_n}=\mathrm{sign}(f(\mathbf x_n))$: the number TP of true positives
where $\widetilde{t_n}=t_n=1$, the number FP of false positives where
$\widetilde{t_n}=1$ but $t_n=-1$, the number TN of true negatives where
$\widetilde{t_n}=t_n=-1$, and the number FN of false negatives where
$\widetilde{t_n}=-1$ but $t_n=1$ (note that the sum of these four numbers is equal
to the number of test data points $|D^{(\mathrm{test})}|$). Given these counts,
one can compute the true positive rate $\text{TPR} =
\text{TP}/(\text{TP}+\text{FN})$ (also known as Recall), the false positive rate
$\text{FPR} = \text{FP}/(\text{FP}+\text{TN})$,  and the
$\text{Precision}=\text{TP}/(\text{TP}+\text{FP})$ (defined to be $1$ if
$\text{TP}+\text{FP}=0$).

Unfortunately, simply using one of these ratios instead of the classification
accuracy does not solve the above problem either. For instance, if we were to measure
success by means of the smallest false positive rate FPR, we would
be satisfied with the trivial all-negative classifier,
since it would never produce a false positive such that $\text{FPR}=0$.

The solution to this kind of problem is to use more robust metrics
such as AUROC (area under
the Receiver Operating Characteristic curve) and AUPRC (area under the
Precision--Recall curve) \cite{Cortes2003AUROCvsErrorRate,
DavisGoadrich2006AUROCandAUPRC}. These metrics are not based on a single evaluation
of the classifier, but rather on the performance of the classifier as a
function of the bias $b$ in \equref{eq:decisionfunction}. By sweeping $b$,
the classifier is artificially moved from an all-negative classifier (corresponding
to $b\to-\infty$, where $\text{TPR}=\text{FPR}=\text{Recall}=0$ and $\text{Precision}=1$)
to an all-positive classifier (corresponding to $b\to\infty$, where $\text{TPR}=\text{FPR}=\text{Recall}=1$).
In essence, this procedure moves the decision boundary through all test data
points, thereby measuring the characteristic shape of the decision boundary.

By plotting TPR vs.~FPR, one generates the ROC curve, and by plotting Precision
vs.~Recall one generates the Precision--Recall curve (see \figref{fig:rocprc}
below for an example of these curves). The area under both curves is termed
AUROC and AUPRC, respectively, and represents a much more robust measure for the
quality of a classifier than the classification accuracy. This means that
optimizing a classifier for AUROC and AUPRC is unlikely to result in a useless
classifier, which can happen when optimizing for the accuracy instead
\cite{Cortes2003AUROCvsErrorRate} (see the example given above).

Note, however, that there is a particular situation in which optimizing for
the accuracy is appropriate, namely to obtain a good value for the bias $b$ in
the decision function given in \equref{eq:decisionfunction}. The reason for this
is that, ultimately, we are interested in making a definite class prediction
$t=\mathrm{sign}(f(\mathbf x))$ for an arbitrary point $\mathbf x$. Since AUROC
and AUPRC are independent of $b$, we cannot use these metrics to obtain
an optimal bias. Instead, a reasonable option is to use the value of $b$ for which the
accuracy with respect to the training data is maximal. This is especially true for
\texttt{qSVM}, for which the candidate given in \equref{eq:biasaveraging} may
not be optimal. This is the case for the real test problem below (see
also \appref{app:bias}).

In the following applications, we report accuracy, AUROC, and AUPRC to compare the
classifiers and to measure the classification performance.

\section{Applications}
\label{sec:applications}

\subsection{Two-dimensional synthetic data}

As a proof of concept and to understand the power of $\texttt{qSVM}$, we
consider a small set of two-dimensional synthetic data. This has the advantage
that the results can be easily visualized and the quality of the many different
classifiers returned by the quantum annealer can be compared.

The dataset $D$ consists of $n=1,\ldots,40$ points $(\mathbf x_n,t_n)$, where
the first half corresponds to the negative class $t_n=-1$ representing an outer
region, and the second half corresponds to the positive class $t_n=1$
representing an inner region. It was generated according to
\begin{align}
  \label{eq:toydatageneration}
  \mathbf x_n &= r_n \begin{pmatrix}
    \cos \varphi_n \\
    \sin \varphi_n
  \end{pmatrix}
  + \begin{pmatrix}
    s_n^x \\
    s_n^y
  \end{pmatrix},
\end{align}
where $r_n = 1$ if $t_n=-1$ and $r_n = 0.15$ if $t_n=1$, $\varphi_n$ is linearly
spaced on $[0,2\pi)$ for each class, and $s_n^x$ and $s_n^y$ are drawn
from a normal distribution with mean $0$ and standard deviation $0.2$.

\begin{figure}
  \centering
  \includegraphics[width=\linewidth]{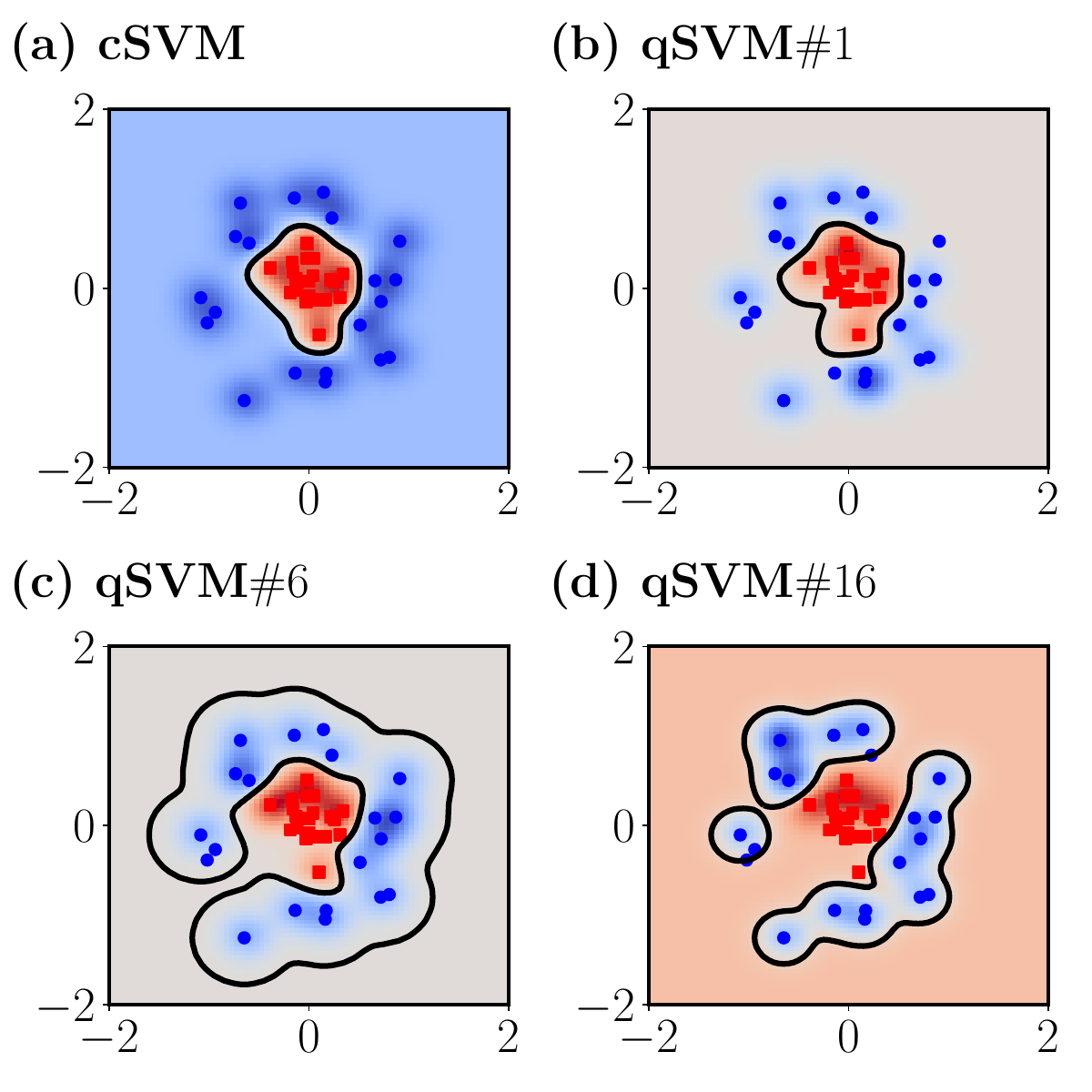}
  \caption{(Color online) Visualization of the classification boundary resulting
  from (a) the global optimum produced by the classical SVM, and (b)--(d)
  various solutions from the ensemble produced by the QA version of the SVM for
  the same problem (the identifier $\texttt{qSVM}\#i$ indicates the
  $(i+1)^{\mathrm{th}}$ sample produced by the DW2000Q, starting at $i=0$ and
  ordered by lowest energy). The parameters for the SVMs are $B=K=2$, $\xi=0$,
  $\gamma=16$, and $C=3$. The two classes for the two-dimensional synthetic data
  are plotted as red squares ($t_n=1$) and blue circles ($t_n=-1$),
  respectively. The corresponding background color indicates the distance to the
  decision boundary.}
  \label{fig:toyproblem}
\end{figure}

We visualize the resulting decision boundaries $f(\mathbf x)=0$ for
$\texttt{cSVM}(3,16)$ in \figref{fig:toyproblem}(a), and for three separate
solutions from the ensemble found by $\texttt{qSVM}(2,2,0,16)$ in
\figref{fig:toyproblem}(b)--(d). For demonstration purposes, the plotted data
points do not come from a separate test set but are the same $40$ points that
the SVM versions have been trained on. The value of the decision  function
$f(\mathbf x)$ given in \equref{eq:decisionfunction} determines the
background color, obtained by evaluating $f(\mathbf x)$ for each point $\mathbf
x$ in the two-dimensional plotting grid.

\begin{figure*}
  \centering
  \includegraphics[width=\linewidth]{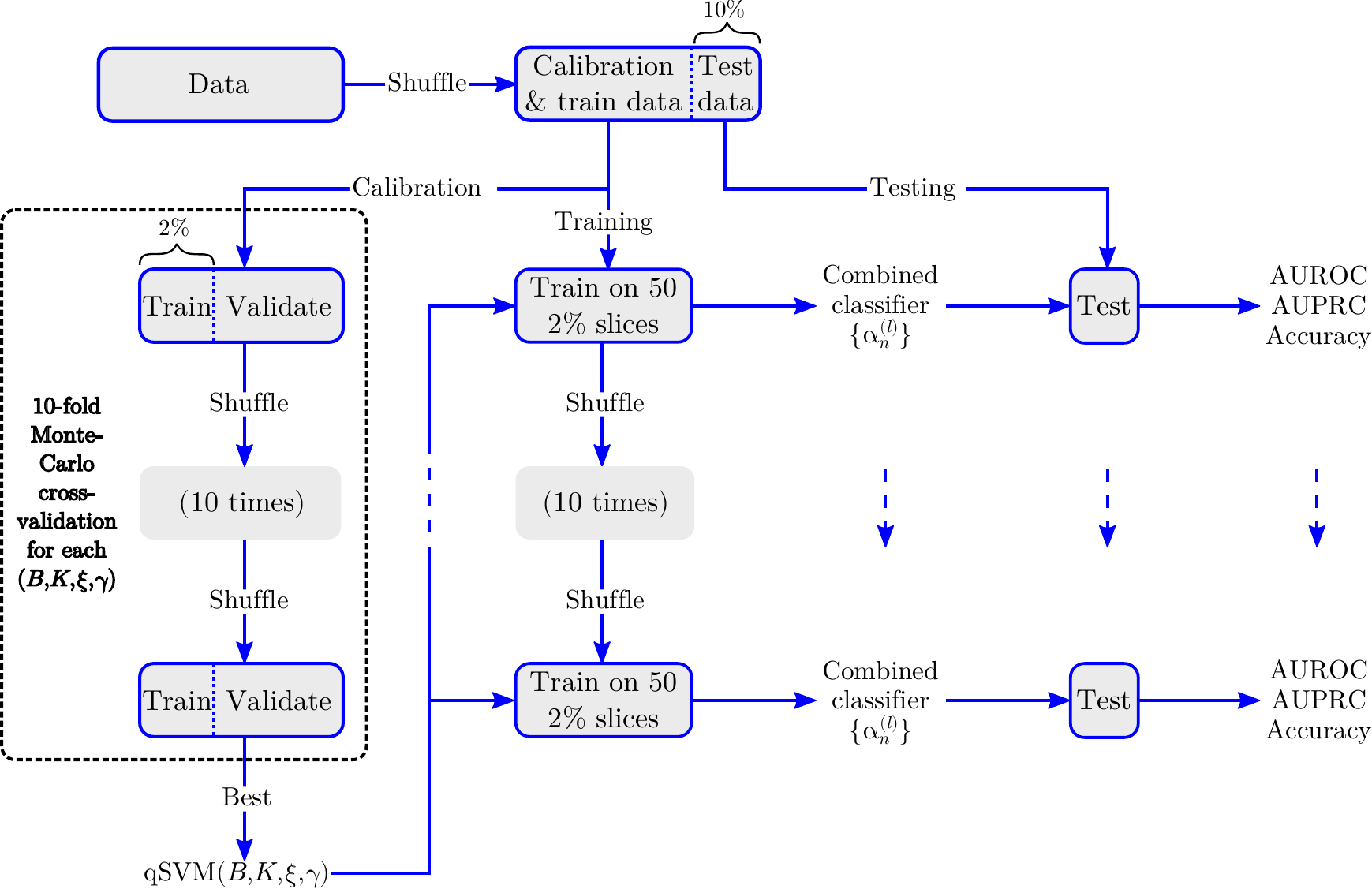}
  \caption{Data handling procedure for the computational biology problem. Each
  of  the nine datasets is split into 90\% calibration and training data
  $D^{(\mathrm{train})}$ and 10\% test data $D^{(\mathrm{test})}$. In the
  calibration phase, 10-fold Monte Carlo  cross-validation is used to select the
  hyperparameters $B$, $K$, $\xi$, and $\gamma$ (see \secref{sec:qSVM}),
  training on 2\% of $D^{(\mathrm{train})}$ and validating on the rest. In the
  test phase, the selected $\texttt{qSVM}(B,K,\xi,\gamma)$ is applied to every
  2\% slice of $D^{(\mathrm{train})}$. The resulting classifiers are combined to
  classify the test data $D^{(\mathrm{test})}$ to evaluate the AUROC, the AUPRC,
  and the classification accuracy (see \secref{sec:evaluation}). The test
  procedure  is repeated 10 times to gather statistics.}
  \label{fig:datahandling}
\end{figure*}

We see that \texttt{cSVM} shown in \figref{fig:toyproblem}(a) satisfies all
the properties expected from the global minimum of an SVM, i.e., separating the
dataset into two regions where the decision boundary has a maximum margin to
the closest data points (the support vectors).

The DW2000Q, however, automatically produces a variety of alternative
classifiers shown in \figref{fig:toyproblem}(b)--(d). Each of them solves the
classification task of the training set as intended, and additionally highlights
different features present in the training data. While sample \#1 shown in
\figref{fig:toyproblem}(b) still resembles the properties of the global minimum,
sample \#6 shown in \figref{fig:toyproblem}(c) yields a more narrow enclosure of
the outer circle. The classifier from sample \#16 shown in
\figref{fig:toyproblem}(d) is even sensitive to the gaps in the outer circle.
This result suggests that a combination of the classifiers returned by
$\texttt{qSVM}$ may be more powerful than the single classifier produced by
$\texttt{cSVM}$.

\subsection{Application to real data}

We compare the performance of both \texttt{cSVM} and \texttt{qSVM} when
applied to real data obtained from the biology experiment studied in
\cite{Li2018TFDNABindingAffinity}. This data was provided to us on request.
Briefly, the classification task is to decide whether a certain protein (a
transcription factor labeled Mad, Max, or Myc) binds to a certain DNA sequence
such as CCCACGTTCT (see also \cite{Zhou2015TFDNABindingAffinity,
Yang2017TFDNAbindingHTSELEX}).

The data consists of nine separate datasets labeled Mad50, Max50, Myc50, Mad70,
Max70, Myc70, Mad80, Max80, and Myc80. The datasets consist of $N=1655$ (Mad),
$N=1599$ (Max), and $N=1584$ (Myc) data points, respectively. The data points
$(\mathbf x_n, t_n)$ for $n=1,\ldots,N$ consist of a 40-dimensional  vector
$\mathbf x_n\in \{-1,+1\}^{40}$ representing the DNA sequence, and a label
indicating whether the protein binds to this DNA sequence ($t_n=+1$) or not
($t_n=-1$). The DNA sequence is encoded by mapping each base-pair in the DNA
alphabet \{A,C,G,T\} according to $\mathrm A\mapsto(+1,-1,-1,-1)$, $\mathrm
C\mapsto(-1,+1,-1,-1)$, $\mathrm G\mapsto(-1,-1,+1,-1)$, and $\mathrm
T\mapsto(-1,-1,-1,+1)$, and concatenating all encoded base-pairs.  An encoding
of this type is sometimes called \emph{one-hot encoding} (often using 0 instead
of $-1$) since only  one element in each encoded base-pair is $+1$ (cf.~also
\cite{Li2018TFDNABindingAffinity, Zhou2015TFDNABindingAffinity}). For each
dataset, the number behind the protein label indicates the percentage of
negative classes such that e.g.~the dataset Max80 contains 80\% non-binding DNA
sequences ($t_n=-1$) and 20\% binding DNA sequences ($t_n=+1$).

We separate each of the nine datasets into 90\% training data $D^{(\mathrm{train})}$
and 10\% test data $D^{(\mathrm{test})}$. The training data is used for calibration
of the hyperparameters and for training the classifiers. The test data is unseen
during training and exclusively used to test the classifiers in the test phase.
The entire data handling procedure is sketched in \figref{fig:datahandling}.

\subsubsection{Calibration phase: Results for a small training dataset}

To select the hyperparameters of $\texttt{qSVM}$, we use 10-fold Monte Carlo (or
split-and-shuffle) cross-validation. This means
that we train $\texttt{qSVM}(B,K,\xi,\gamma)$ on 2\% of $D^{(\mathrm{train})}$
(approximately 30 data points) and evaluate its performance on the remaining
data points of $D^{(\mathrm{train})}$ for validation. The data is then shuffled
and the process is repeated a total number of ten times (see \figref{fig:datahandling}).

\begin{figure}
  \centering
  \includegraphics[width=\linewidth]{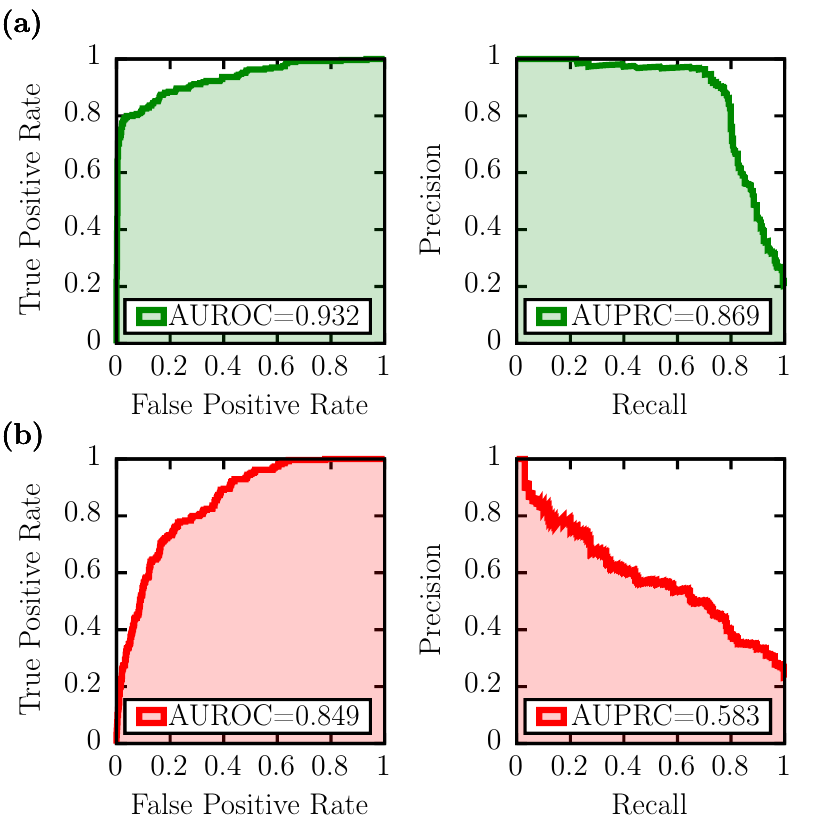}
  \caption{(Color online) Example for the generated ROC and PR curves to measure
  the quality of the classifiers. (a) $\texttt{qSVM}(10,3,0,-1)\#14$ using
  $n_{\mathrm{cpl}}=2000$ couplers, and (b) $\texttt{cSVM}(111,-1)$  (note that
  $C=111$ for \texttt{cSVM} corresponds to $B=10$ and $K=3$, see
  \equref{eq:boxconstraintparameter}).  Both SVMs have been trained and
  validated on the same data, taken from the fifth step in the 10-fold
  cross-validation procedure  for the dataset Max80
  \cite{Li2018TFDNABindingAffinity}.}
  \label{fig:rocprc}
\end{figure}

The small fraction of 2\% was chosen because of the size limitations  of the
quantum annealer (cf.~also \cite{Li2018TFDNABindingAffinity}).  Since this
is a  very small amount of data, we performed some initial tests before
systematically  calibrating the hyperparameters. In these tests, we observed
that $\texttt{qSVM}$ can produce significantly stronger classifiers than
$\texttt{cSVM}$ for the same little training data and parameters. One example is
shown in \figref{fig:rocprc}, where the ROC and PR curves are plotted for
$\texttt{qSVM}(10,3,0,-1)\#14$ (see \figref{fig:rocprc}(a)) and for
$\texttt{cSVM}(111,-1)$ (see \figref{fig:rocprc}(b)), generated by sweeping the
bias $b$ as explained in \secref{sec:evaluation}. While the QA version produces
almost optimal curves, the global optimum from the classical SVM obviously lacks
precision when applied to the much larger validation data.

For each dataset, the hyperparameters are calibrated by evaluating
$\texttt{qSVM}$ for $B\in\{2,3,5,10\}$ and $K\in\{2,3\}$
(cf.~\equref{eq:encoding}), $\xi\in\{0,1,5\}$ (cf.~\equref{eq:qsvmfull}), and
$\gamma\in\{-1,0.125,0.25,0.5,1,2,4,8\}$ (cf.~\equref{eq:kernel}). We
generically consider the classifiers $\{\alpha_n^{(i)}\}$ from the twenty best
solutions $\texttt{qSVM}(B,K,\xi,\gamma)\#i$ for $i=0,\ldots,19$ as described in
\secref{sec:qSVM}. The evaluation is repeated ten times for the Monte Carlo
cross-validation. Therefore, each set of hyperparameters for each dataset
results in a total of 200 values for AUROC, AUPRC, and accuracy.

\begin{figure}
  \centering
  \includegraphics[width=\linewidth]{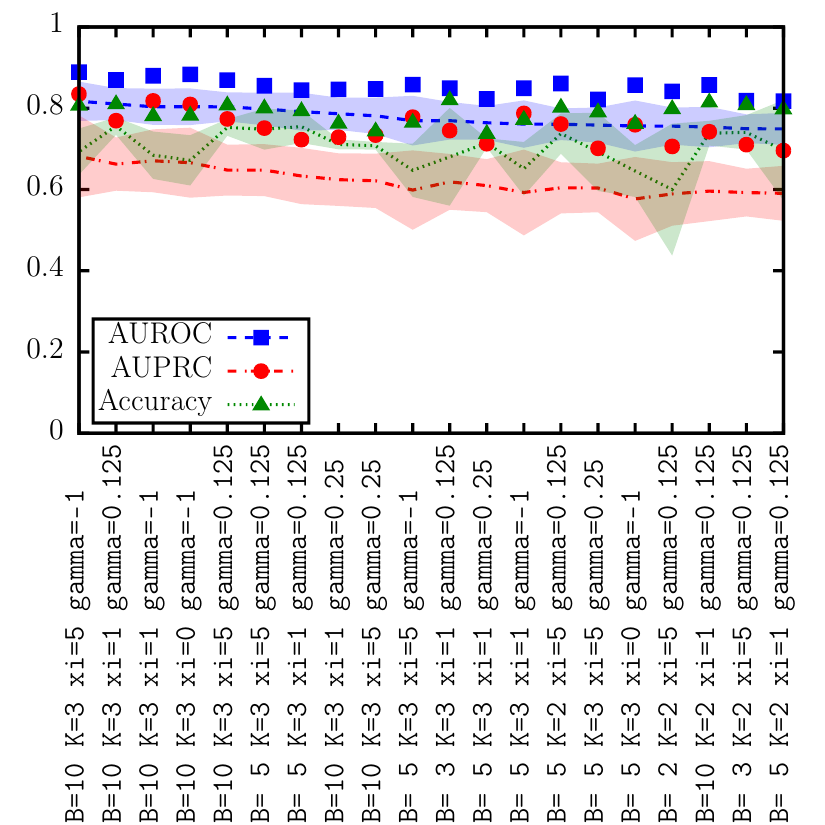}
  \caption{(Color online) Calibration performance of \texttt{qSVM} for the best
  sets of hyperparameters $(B,K,\xi,\gamma)$, ordered by mean AUROC, for the
  dataset Max70 \cite{Li2018TFDNABindingAffinity}. Shown are the AUROC (blue
  dashed line), the AUPRC (red dash-dotted line), the accuracy (green dotted
  line), and the respective standard deviations (shaded areas) over 200
  classifiers (10 different calibration folds times 20 of the best solutions
  from the DW2000Q). Lines connecting the averages are guides to the eye.
  Squares, circles, and triangles denote the maximum performance among each of
  the 200 classifiers.}
  \label{fig:calibration}
\end{figure}

An example of the calibration procedure for the dataset Max70 is shown in
\figref{fig:calibration}. For this dataset, we see that the linear kernels
denoted by $\gamma=-1$ (see \equref{eq:kernel}) dominate (but Gaussian kernels
perform still reasonably well). The selected set of hyperparameters in this case
is $B=10$, $K=3$, $\xi=5$, and $\gamma=-1$, corresponding to the leftmost points
in \figref{fig:calibration}. We also see fluctuations in the mean accuracy
which are not reflected by AUROC and AUPRC. Since AUROC and AUPRC are
insensitive to the  bias, this indicates that the choice for the bias $b$ given
by \equref{eq:biasaveraging} may not always be optimal (see  \appref{app:bias}
for a way to improve the bias if the accuracy matters).

We selected the hyperparameters based on both mean AUROC and AUPRC. The reason
for this is that we observed, when selecting exclusively based on the best AUPRC
(cf.~\cite{Li2018TFDNABindingAffinity}), we sometimes obtained hyperparameters
yielding $\mathrm{AUROC}\approx0.5$ (the result for a random classifier
\cite{Cortes2003AUROCvsErrorRate}).

\begin{table}
  \caption{\label{tab:calibration} Selected hyperparameters for each dataset
  \cite{Li2018TFDNABindingAffinity}. The parameters are the base $B$, the number
  $K$ of qubits per coefficient $\alpha_n$, the  multiplier $\xi$, the kernel
  parameter $\gamma$, and the box constraint parameter $C$ (see
  \secref{sec:SVM}). The value of $C$ is fixed by $B$ and $K$ through
  \equref{eq:boxconstraintparameter} and is given for reference only.}
\begin{ruledtabular}
\begin{tabular}{@{}lccccc@{}}
 Dataset & $B$ & $K$ & $\xi$ & $\gamma$ & $C$\\
  \colrule
  {Mad50} & $2$ & $3$ & $5$ & $0.125$ & $7$ \\
  {Max50} & $2$ & $3$ & $5$ & $0.125$ & $7$ \\
  {Myc50} & $2$ & $2$ & $0$ & $0.125$ & $3$ \\
  {Mad70} & $10$ & $3$ & $5$ & $-1$ & $111$ \\
  {Max70} & $10$ & $3$ & $5$ & $-1$ & $111$ \\
  {Myc70} & $10$ & $3$ & $5$ & $-1$ & $111$ \\
  {Mad80} & $10$ & $3$ & $5$ & $-1$ & $111$ \\
  {Max80} & $10$ & $3$ & $0$ & $-1$ & $111$ \\
  {Myc80} & $10$ & $3$ & $5$ & $-1$ & $111$ \\
\end{tabular}
\end{ruledtabular}
\end{table}

In \tabref{tab:calibration}, we list the best hyperparameters selected for each
dataset. The trend from Gaussian kernels to linear kernels  can be observed in
all datasets: For Mad50, Max50, and Myc50, where half of the data is classified
as  positive and the other half as negative, only the Gaussian kernels can
produce a  reasonable decision boundary (see also \tabref{tab:results} in
\appref{app:numericalresults}). But when going to higher class imbalances as
present in the datasets Mad80, Max80, and Myc80,  a linear decision boundary
suffices to classify the DNA sequences.

The numerical results of the calibration procedure for each dataset in
comparison with the corresponding \texttt{cSVM} are listed in
\tabref{tab:results} in \appref{app:numericalresults}.

\subsubsection{Training and test phase: Results for a larger training dataset}

In this section, we examine a way to overcome the size limitations of the
DW2000Q for real applications with a bigger training dataset. We take the same
nine DNA datasets as before, but now consider the full datasets
$D^{(\mathrm{train})}$ for training a classifier. The goal is to construct an
aggregated classifier from the results of $\texttt{qSVM}$ trained on each $2\%$
slice of the available training data (see Fig. \figref{fig:datahandling}). Each
of the $L=50$ slices is labeled $D^{(\mathrm{train},l)}$ for $l=0,\ldots,49$.
The hyperparameters for each dataset are taken from the calibration results
listed in \tabref{tab:calibration}.

The combined classifier is constructed in two steps. First, for each slice
$D^{(\mathrm{train},l)}$, the twenty best solutions from the DW2000Q (labeled
$\texttt{qSVM}(B,K,\xi,\gamma)\#i$ for $i=0,\ldots,19$) are combined by
averaging over the respective decision functions $f^{(l,i)}(\mathbf x)$ (see
\equref{eq:decisionfunction}). Since the decision function is linear in the
coefficients and the bias ($b^{(l,i)}$ is computed from $\alpha_n^{(l,i)}$ via
\equref{eq:biasaveraging}), this procedure effectively results in one classifier
with an effective set of coefficients $\alpha_n^{(l)} = \sum_i
\alpha_n^{(l,i)} / 20$ and an effective bias $b^{(l)} = \sum_i b^{(l,i)}/20$.

The second step is to average over the $L=50$ slices. Note, however, that the
data points $(\mathbf x_n^{(l)},t_n^{(l)})\in D^{(\mathrm{train},l)}$ are now
different for each $l$. The full decision function is
\begin{align}
  \label{eq:fulldecisionfunction}
  F(\mathbf x) = \frac 1 L \sum\limits_{nl} \alpha_n^{(l)} t_n^{(l)}
  k(\mathbf x_n^{(l)},\mathbf x) + b,
\end{align}
where $b=\sum_{l}b^{(l)}/L$. As before, a decision for the class label of
a point $\mathbf x$ is obtained through $\widetilde{t} = \mathrm{sign}(F(\mathbf
x))$. We use this decision function to evaluate the metrics discussed in
\secref{sec:evaluation} for the test data $D^{(\mathrm{test})}$ using the
procedure  illustrated in \figref{fig:datahandling}.

Note that in \cite{Li2018TFDNABindingAffinity}, instead of separating the
training data into 50 disjoint subsets (each containing 2\% of the data), an
approach similar to bagging (bootstrap aggregating) \cite{Breiman1996Bagging}
was used. In that approach, 50 subsets are constructed by drawing $2\%$ of the
training data \emph{with replacement}. We also tested this bagging inspired
approach (data not shown) and found that, although the results were similar, the
fluctuations were much larger. This makes sense because drawing with
replacement means that different subsets can share the same data points and
also include a single point more than once. Consequently, one may expect that
some points are not included in any of the datasets. In fact, the probability
that a certain $x\in D^{(\mathrm{train})}$ is not included in any of the
$D^{(\mathrm{train},l)}$ is $(1-1/N)^N\approx 36.8\%$ for
$N=|D^{(\mathrm{train})}|\approx1500$. Apart from this counting argument, the
general observation in \cite{Breiman1996Bagging} was that bagging is better
suited for \emph{unstable} classification algorithms, whereas SVMs are stable.
We therefore conclude that splitting the training data in disjoint,
equally-sized subsets is superior.

\begin{figure}
  \centering
  \includegraphics[width=\linewidth]{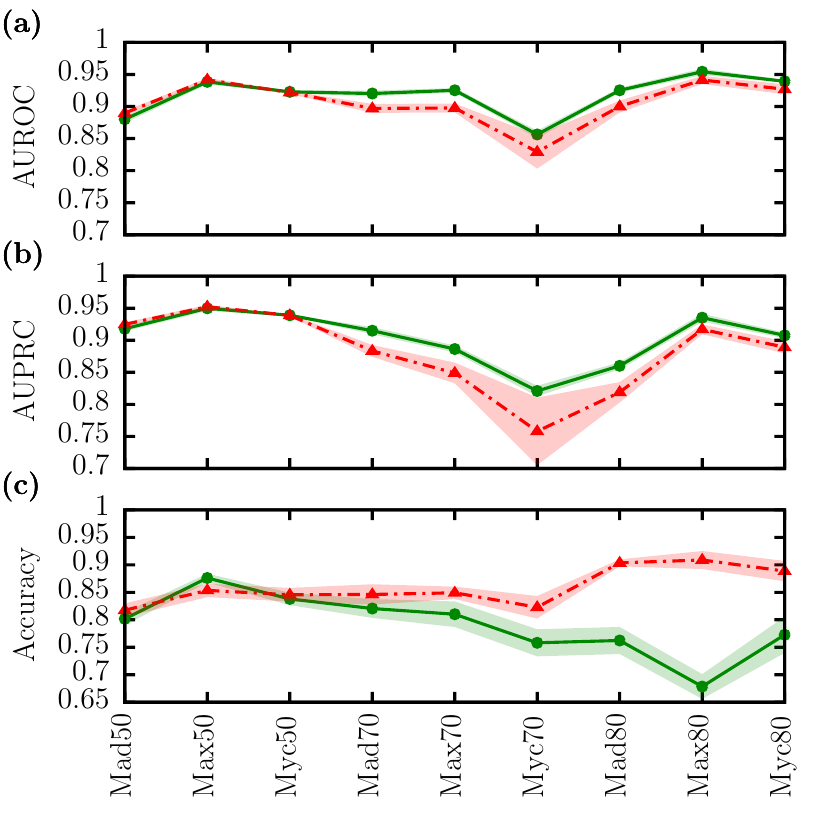}
  \caption{(Color online) Performance of $\texttt{qSVM}$ (solid green line) and
  $\texttt{cSVM}$ (dash-dotted red line) as measured by (a) AUROC, (b) AUPRC,
  and (c) accuracy (see \secref{sec:evaluation}) using the decision function
  given in \equref{eq:fulldecisionfunction} for each of the nine datasets from
  the computational biology problem \cite{Li2018TFDNABindingAffinity}. The
  parameters for each dataset are taken from \tabref{tab:calibration}. The
  standard deviation over ten repetitions (see \figref{fig:datahandling}) is
  shown as shaded areas. Lines are guides to the eye.}
  \label{fig:finaltraining}
\end{figure}

As before, it is interesting to compare the results from the combined classifier
with results from applying \texttt{cSVM} to the same data points and parameters.
Note that \equref{eq:fulldecisionfunction} also applies to \texttt{cSVM}, but
that $\alpha_n^{(l)}$ comes directly from the global minimum to
\equsref{eq:qp}{eq:qpzero} and not from an average of the twenty best solutions
produced by DW2000Q. The results for each dataset are shown in
\figref{fig:finaltraining}, where the mean and the standard deviation have been
obtained from ten repetitions of the test procedure as sketched in
\figref{fig:datahandling}.

\begin{figure}
  \centering
  \includegraphics[width=\linewidth]{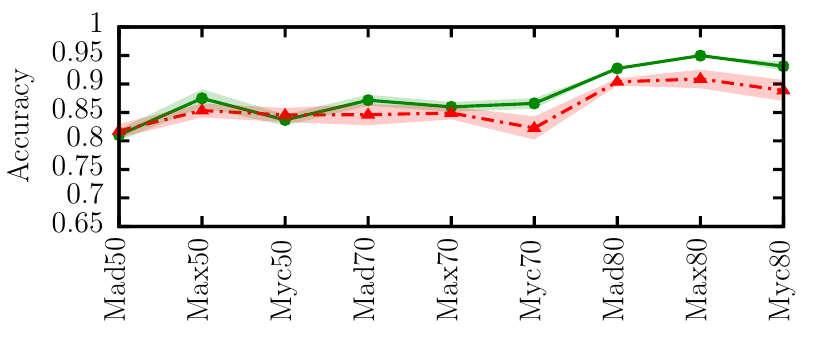}
  \caption{(Color online) Classification accuracy of $\texttt{qSVM}$ (solid
  green line) and $\texttt{cSVM}$ (dash-dotted red line) as shown in
  \figref{fig:finaltraining}(c), after adjusting the suboptimal bias $b$ to
  $b^*$ where the accuracy for the training data is higher (see
  \appref{app:bias}). The metrics AUROC and AUPRC are the same as in
  \figref{fig:finaltraining}(a) and (b).}
  \label{fig:finaltrainingheuristic}
\end{figure}

Based on the resulting accuracy shown in \figref{fig:finaltraining}(c), one
could conclude that \texttt{cSVM} outperforms $\texttt{qSVM}$ (especially
for the dataset Max80 for which we studied one of the contributing classifiers
in \figref{fig:rocprc}). However, from the metrics AUROC and AUPRC reported in
\figref{fig:finaltraining}(a) and (b), we find that the resulting classifiers
from the QA version are in fact superior. This hints at a problem in the
construction of the final decision function given in
\equref{eq:fulldecisionfunction}, which would have been overlooked if the
accuracy had not been evaluated: Recall that AUROC and AUPRC are generated by
sweeping the bias $b$ in \equref{eq:fulldecisionfunction}  to move the decision
boundary through the feature space $\mathbb R^{40}$ from a full negative
predictor to a fully positive predictor (see \secref{sec:evaluation}). If  AUROC
and AUPRC are better for $\texttt{qSVM}$, this means that the bias $b$  has been
chosen suboptimally and there must be some bias $b^*$ for which the classifier
produces better results.

The reason for this is that \equref{eq:biasaveraging} from the original SVM may
not be suited to obtain the optimal bias for the QA version of the SVM
defined by \equref{eq:qsvmfull}. The condition for an optimal bias is the
constraint \equref{eq:qpzero}, included through the  multiplier $\xi$ in
\equref{eq:qsvmfull}. Since $\xi=0$ for Max80 (cf.~\tabref{tab:calibration}),
this explains the particularly bad accuracy for this dataset despite better
AUROC and AUPRC (see also the discussion under point 6 of the motivations given
in \secref{sec:qSVM}).

We correct for the suboptimal bias by replacing $b$ with the value of $b^*$ for
which the classification accuracy for the training data $D^{(\mathrm{train})}$ is maximal.
Note that this step does not require a new training of \texttt{qSVM}. It is
a scan of a single parameter that can be done ``offline'', i.e., after obtaining
the coefficients $\{\alpha_n^{(l)}\}$. As such, this step is simple and
efficient and could, in principle, be directly added to the data handling scheme
presented in \figref{fig:datahandling}. See \appref{app:bias} for more
information.

The classification accuracy of $\texttt{qSVM}$ after adjusting the bias for
each dataset is shown in \figref{fig:finaltrainingheuristic}. It clearly
improves the results for the linear kernel ($\gamma=-1$) with high class imbalance
(Mad80, Max80, and Myc80). We also observe that the Gaussian kernel used for
Mad50, Max50, and Myc50 was not affected as strongly by the suboptimal bias.
As changing the bias of the decision function given in
\equref{eq:fulldecisionfunction}  does not affect AUROC and AUPRC, the results
shown in \figref{fig:finaltraining}(a) and (b) also apply to the adjusted
version of $\texttt{qSVM}$.

To summarize, we observe a better or comparative performance of
$\texttt{qSVM}$ compared to \texttt{cSVM} for all datasets, as measured by
AUROC, AUPRC, and classification accuracy. For completeness, the numerical
results of the test are given in \tabref{tab:results} in
\appref{app:numericalresults}.

\section{Conclusion}
\label{sec:conclusion}

In this paper, we introduced and studied the implementation of kernel-based SVMs
on a DW2000Q quantum annealer \cite{DW2000Q}. We found that the optimization
problem behind the training of SVMs can be straightforwardly expressed as a QUBO
and solved on a quantum annealer. The QUBO form exhibits certain mathematical
advantages, such as its ability to produce exact zeros or the inherent inclusion
of the box constraint. Each run of the training process on the quantum annealer
yields a distribution of different classifiers that can later be used to
classify arbitrarily many test data points.

Our results show that the ensemble of classifiers produced by the quantum
annealer often surpasses the single classifier obtained by the classical SVM for
the same computational problem as measured by AUROC, AUPRC, and accuracy. The
advantage stems from the fact that the DW2000Q produces not just the global
optimum for the training data, but a distribution of many reasonably good,
close-to-optimal solutions to the given optimization problem. A combination of
these has the potential to generalize better to the test data. This observation
is in line with findings in other machine learning problems studied on a quantum
annealer \cite{Mott2017HiggsQML, Li2018TFDNABindingAffinity}.

Therefore we conclude that the QA version of the SVM is a useful practical
alternative to the classical SVM. If the capabilities of future quantum
annealers continue to scale at the current pace, training SVMs on quantum
annealers may become a valuable tool for classification problems, and can
already be helpful for hard problems where only little training data is
available.

An interesting project for future research would be to examine other approaches
to building strong classifiers by constructing weighted sums of the class
predictions from several SVMs as done in boosting methods like AdaBoost or
QBoost \cite{Neven2012QBoost, Freund1996AdaBoost,
Friedman2000StatisticalViewBoosting, Bishop2006MachineLearning}.  It would also
be valuable to examine how the QA results for SVMs can be further improved using
advanced features offered by the DW2000Q like reverse annealing,
spin-reversal transforms, special annealing schedules, or enhanced
embeddings \cite{Ohkuwa2018DWaveReverseAnnealing,
Boixo2013ExperimentalSignatureQA, DW2000Q}. Furthermore, since SVMs can also be
used for multi-class classification and regression tasks
\cite{schoelkopf2002learningwithkernels}, it seems worthwhile to study
corresponding applications to such problems using the QA formulation presented
here. Finally, it would be a potentially interesting avenue to explore if
suitable modifications to the original SVM can lead to an equally good
distribution of solutions as the one produced by the quantum annealer.

\section{Acknowledgments}

We would like to thank Richard Li and Daniel Lidar for providing preprocessed
data from TF-DNA binding experiments. We are grateful to Seiji Miyashita for
helpful discussions. Access and compute time on the D-Wave machine located at
the headquarters of D-Wave Systems Inc.~in Burnaby (Canada) were provided by
D-Wave Systems Inc. D.W. is supported by the Initiative and Networking Fund of
the Helmholtz Association through the Strategic Future Field of Research project
``Scalable solid state quantum computing (ZT-0013).''

\bibliographystyle{apsrev4-1custom}
\bibliography{bibliography}

\appendix

\FloatBarrier
\onecolumngrid

\section{Adjusting the bias in \texttt{qSVM}}
\label{app:bias}

The choice for the bias $b$ given in \equref{eq:biasaveraging} as a
function of the coefficients $\{\alpha_n\}$ is based on the condition that the
coefficients are the global minimum $\{\alpha_n\}^*$ of the QP problem given in
\equsref{eq:qp}{eq:qpzero}. In fact, it is the constraint given in
\equref{eq:qpzero} that identifies an optimal bias $b$ \cite{numericalrecipes}.

However, for \texttt{qSVM}, a new classifier is generated by combining  some of
the lowest-energy solutions produced by the quantum annealer, which  is in
general not equal to $\{\alpha_n\}^*$. Moreover,  the constraint for an optimal
bias given in \equref{eq:qpzero} is included through  the  multiplier
$\xi$ in \equref{eq:qsvmfull}, so it may not be  satisfied for all solutions
produced by the quantum annealer. Therefore, it  can happen that the bias from
\equref{eq:biasaveraging} is not suitable for  \texttt{qSVM}. This is what
happened to the rightmost three datasets shown  in \figref{fig:finaltraining}
(especially for Max80 where $\xi=0$, see \tabref{tab:calibration}).  This
problem only affects the actual accuracy and not the more robust metrics AUROC
and AUPRC (see \secref{sec:evaluation}).

Since the bias is only a single parameter, this problem can easily be solved by
replacing $b$ with another bias $b^*$, for which the accuracy for the training
data $D^{(\mathrm{train})}$ is highest. Note that this step does not require
a new training of \texttt{qSVM}. The result of the training is given by
the set of coefficients $\{\alpha_n\}$, which enter the decision function in
\equref{eq:decisionfunction}. The bias $b$ can be adjusted ``offline'', i.e.,
independently of the $\{\alpha_n\}$.

Note that it is only allowed to use the training data $D^{(\mathrm{train})}$ for
adjusting the bias, and not the test data $D^{(\mathrm{test})}$. Modifying a
classifier as a function of its  performance on the test data
$D^{(\mathrm{test})}$ would invalidate the statement that the classifier can
generalize well to unseen data.

\begin{figure}[h]
  \centering
  \includegraphics[width=.8\linewidth]{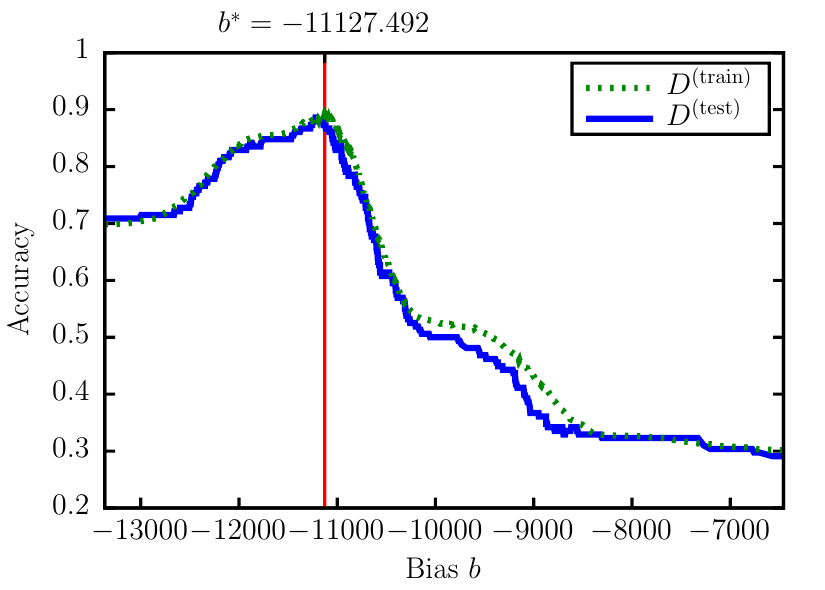}
  \caption{(Color online) Classification accuracy for the training data
  $D^{(\mathrm{train})}$ (dotted green line) and the test data
  $D^{(\mathrm{test})}$ (solid blue line) of the dataset Myc70 as a function
  of the bias $b$ in the decision function $F(\mathbf x)$ given in
  \equref{eq:fulldecisionfunction}. The bias $b^*$ is chosen to be
  optimal for the training data. The optimal bias for the test data (i.e. the
  peak of the solid blue line) is slightly smaller.}
  \label{fig:finaltrainingheuristicscanb}
\end{figure}

An example of such a scan of the bias $b$ is shown in
\figref{fig:finaltrainingheuristicscanb} for the dataset Myc70. It was taken
from one out of ten repetitions of the test procedure (see
\figref{fig:datahandling}). The classifier has been obtained from an average
over 1000 decision functions  (20 lowest-energy samples times 50 slices of the
training data). One can see that the peak of the accuracy for
$D^{(\mathrm{train})}$ (dotted line) is close but not equal to the peak of the
accuracy for $D^{(\mathrm{test})}$ (solid line).

\clearpage

\section{Calibration and test results}
\label{app:numericalresults}

In \tabref{tab:results}, we list the numerical results for the calibration and
the test phase for the application of \texttt{cSVM} and \texttt{qSVM} to the
computational biology problem.

For the calibration phase, where  2\% of the data was used for training,
$\texttt{qSVM}$ often produces stronger or equally strong classifiers. In the
testing phase, where the classifiers for each of the 50 disjoint subsets of the
training data were combined,  $\texttt{qSVM}$ almost always surpasses
\texttt{cSVM} in all of the three metrics.

\begin{table*}[h]
  \caption{\label{tab:results} Calibration and test results for all SVMs. The
  reported metrics are the mean area under the ROC curve, and the mean area
  under the  Precision--Recall curve (see \secref{sec:evaluation}), and  the mean
  classification accuracy. The parameters of the QA version of the SVM  are
  $\texttt{qSVM}(B,K,\xi,\gamma)$ where $B$ is the encoding base, $K$ is the
  number of  qubits per coefficient $\alpha_n$, $\xi$ is a Lagrangian
  multiplier, and $\gamma$ is the kernel parameter.  The corresponding version
  of the classical SVM is $\texttt{cSVM}(C,\gamma)$ where $C$ is given by
  \equref{eq:boxconstraintparameter}.}
\begin{ruledtabular}
\begin{tabular}{@{}llcccccc@{}}
 \multirow{2}{*}{Dataset} & \multirow{2}{*}{SVM Parameters} & \multicolumn{3}{c}{Calibration} & \multicolumn{3}{c}{Testing} \\ \cline{3-5}\cline{6-8}
  && AUROC & AUPRC & Accuracy & AUROC & AUPRC & Accuracy \\
  \colrule
  \multirow[c]{2}{*}{Mad50} & \texttt{qSVM(2,3,5,0.125)} & $0.71$ & $0.71$ & $0.63$ & $0.88$ & $0.92$ & $0.81$ \\
                            & \texttt{cSVM(7,0.125)} & $0.73$ & $0.73$ & $0.60$ & $0.89$ & $0.92$ & $0.82$ \\
  \colrule
  \multirow[c]{2}{*}{Max50} & \texttt{qSVM(2,3,5,0.125)} & $0.73$ & $0.74$ & $0.64$ & $0.94$ & $0.95$ & $0.87$ \\
                            & \texttt{cSVM(7,0.125)} & $0.73$ & $0.74$ & $0.63$ & $0.94$ & $0.95$ & $0.85$ \\
  \colrule
  \multirow[c]{2}{*}{Myc50} & \texttt{qSVM(2,2,0,0.125)} & $0.68$ & $0.68$ & $0.61$ & $0.92$ & $0.94$ & $0.84$ \\
                            & \texttt{cSVM(3,0.125)} & $0.69$ & $0.70$ & $0.58$ & $0.92$ & $0.94$ & $0.85$ \\
  \colrule
  \multirow[c]{2}{*}{Mad70} & \texttt{qSVM(10,3,5,-1)} & $0.75$ & $0.58$ & $0.65$ & $0.92$ & $0.91$ & $0.87$ \\
                            & \texttt{cSVM(111,-1)} & $0.70$ & $0.47$ & $0.67$ & $0.90$ & $0.88$ & $0.85$ \\
  \colrule
  \multirow[c]{2}{*}{Max70} & \texttt{qSVM(10,3,5,-1)} & $0.82$ & $0.68$ & $0.69$ & $0.93$ & $0.89$ & $0.86$ \\
                            & \texttt{cSVM(111,-1)} & $0.75$ & $0.57$ & $0.70$ & $0.90$ & $0.85$ & $0.85$ \\
  \colrule
  \multirow[c]{2}{*}{Myc70} & \texttt{qSVM(10,3,5,-1)} & $0.72$ & $0.57$ & $0.63$ & $0.86$ & $0.82$ & $0.87$ \\
                            & \texttt{cSVM(111,-1)} & $0.72$ & $0.51$ & $0.66$ & $0.83$ & $0.76$ & $0.82$ \\
  \colrule
  \multirow[c]{2}{*}{Mad80} & \texttt{qSVM(10,3,5,-1)} & $0.85$ & $0.66$ & $0.69$ & $0.93$ & $0.86$ & $0.93$ \\
                            & \texttt{cSVM(111,-1)} & $0.78$ & $0.50$ & $0.78$ & $0.90$ & $0.82$ & $0.90$ \\
  \colrule
  \multirow[c]{2}{*}{Max80} & \texttt{qSVM(10,3,0,-1)} & $0.85$ & $0.62$ & $0.67$ & $0.95$ & $0.94$ & $0.95$ \\
                            & \texttt{cSVM(111,-1)} & $0.78$ & $0.47$ & $0.77$ & $0.94$ & $0.92$ & $0.91$ \\
  \colrule
  \multirow[c]{2}{*}{Myc80} & \texttt{qSVM(10,3,5,-1)} & $0.73$ & $0.48$ & $0.60$ & $0.94$ & $0.91$ & $0.93$ \\
                            & \texttt{cSVM(111,-1)} & $0.71$ & $0.37$ & $0.71$ & $0.93$ & $0.89$ & $0.89$ \\
\end{tabular}
\end{ruledtabular}
\end{table*}

\end{document}